\documentclass[preprint]{neus2025}

\title{Logic Gate Neural Networks are Good for Verification}
\usepackage{times}
\usepackage{booktabs}
\usepackage{multirow}
\usepackage{comment}
\usepackage{mathtools}
\usepackage{wrapfig}
\usepackage{todonotes}
\usepackage{tablefootnote}
\usepackage{subcaption}
\usepackage{caption}
\usepackage{graphicx}
\usepackage{pgfplots}
\usepackage{xstring}
\pgfplotsset{compat=1.18}
\usepackage{pgfplotstable}
\usepgfplotslibrary{groupplots}

\coltauthor{\Name{Fabian Kresse} \Email{fabian.kresse@ist.ac.at}\\
 \Name{Emily Yu} \Email{emily.yu@ist.ac.at}\\
 \Name{Christoph H. Lampert} \Email{chl@ist.ac.at}\\
\Name{Thomas A. Henzinger} \Email{tah@ist.ac.at}\\
 \addr Institute of Science and Technology Austria (ISTA), Klosterneuburg, Austria}

\begin{document}

\maketitle
\pagestyle{plain}
\thispagestyle{plain}
\begin{abstract}
Learning-based systems are increasingly deployed across various domains, yet the complexity of traditional neural networks poses significant challenges for formal verification. Unlike conventional neural networks, learned Logic Gate Networks (LGNs) replace multiplications with Boolean logic gates, yielding a sparse, netlist-like architecture that is inherently more amenable to symbolic verification, while still delivering promising performance. In this paper, we introduce a SAT encoding for verifying global robustness and fairness in LGNs. 
We evaluate our method on five benchmark datasets, including a newly constructed 5-class variant, and find that LGNs are both verification-friendly and maintain strong predictive performance.
\end{abstract}


\section{Introduction}
Neural networks are increasingly utilized in emerging safety-critical domains such as self-driving cars, robotic systems, and healthcare devices, as they show great promise in solving complex real-world tasks~\citep{jumper2021highly,fawzi2022discovering}. However, ensuring correctness and reliability in these %
settings poses significant challenges, driven by the networks’ inherent opacity and high complexity~\citep{AmodeiOSCSM16}. As a result, numerous verification methods have been put forward to address these issues by formally proving the robustness and fairness of neural networks. Among the commonly considered properties is local robustness, such that for a given input, a network is robust to a set of specified perturbations~\citep{DBLP:conf/cav/KatzBDJK17,DBLP:conf/sp/GehrMDTCV18,DBLP:conf/nips/SinghGMPV18}. Another critical property is fairness, which requires that the network’s predictions be free from bias with respect to attributes such as gender, ethnicity, and age. Extending these notions further, global robustness and fairness are 2-safety hyperproperties demanding that a network maintain robustness and fairness for all pairs of inputs \citep{DBLP:conf/cav/AthavaleBCMNW24,DBLP:journals/pacmpl/KabahaD24, DBLP:conf/icse/BiswasR23,DBLP:conf/aaai/KhedrS23}. 

Verification methods for neural networks can be grouped into two main categories: complete and incomplete. Incomplete approaches often employ semidefinite programming~\citep{DBLP:conf/nips/RaghunathanSL18,DBLP:conf/nips/DathathriDKRUBS20,DBLP:journals/tac/FazlyabMP22} or bound-propagation techniques~\citep{DBLP:conf/nips/WangPWYJ18,DBLP:conf/uss/WangPWYJ18,DBLP:conf/nips/WangZXLJHK21,DBLP:journals/pacmpl/SinghGPV19} to overapproximate the network’s behavior. Although this yields efficient verification in practice, it may introduce approximation errors. In contrast, complete methods use mixed-integer linear programming (MILP)~\citep{DBLP:conf/iclr/TjengXT19,DBLP:conf/ipco/AndersonHTV19} and satisfiability modulo theories (SMT)~\citep{DBLP:conf/cav/WuIZTDKRAJBHLWZKKB24,DBLP:journals/aicom/PulinaT12, katz2019marabou} to provide definitive verdicts to verification queries by encoding the neural networks. However, the exhaustive nature of these techniques often results in higher computational overhead, particularly for large-scale deep neural networks (DNNs).

Despite rapid advances in DNN verification, most efforts have focused on conventional neural networks, for which the verification task still faces significant challenges. As an alternative, some methods use quantized networks~\citep{DBLP:journals/corr/abs-2103-13630,DBLP:conf/aaai/LechnerZCHR23,DBLP:conf/aaai/HenzingerLZ21} and binary neural networks~\citep{DBLP:journals/pr/QinGLBSS20}, whose verification can be performed using SMT/SAT solvers~\citep{DBLP:conf/nips/JiaR20}.
Recently, deep differentiable Logic Gate Networks (LGNs) have attracted attention for their potential to deliver fast, efficient inference while exhibiting promising performance~\citep{petersen2022deep,petersen2024convolutional}. Unlike binary neural networks, which discretize weights and activations, LGNs use a differentiable relaxation to learn a mixture over basic logic gate operators (e.g., NAND and XOR). LGNs can be directly deployed on FPGAs or ASICs, potentially yielding significant energy savings over conventional architectures and making them especially attractive for cyber-physical systems. 
Furthermore, due to their discrete and SAT-solving-friendly architecture, LGNs and related variants have shown promise for formal verification, particularly in the context of local robustness~\citep{DBLP:conf/ijcai/BenamiraPYGH24}.

In this paper, we take a first step toward SAT-based verification of learned logic gate networks by focusing on two global properties: robustness and fairness. 
Global robustness and fairness say that similar inputs should yield similar results;
by comparing two input-output pairs,
they are 2-safety hyperproperties~\citep{DBLP:conf/cav/AthavaleBCMNW24}. 
Specifically, we propose a symbolic encoding that formalizes these hyperproperties in a confidence-based framework, enabling efficient verification by a SAT solver. We then conduct an experimental evaluation on five datasets, illustrating the promise of our approach for the practical verifiability and the competitive predictive performance of LGNs. To the best of our knowledge, this is the first exploration of using SAT solving for 2-safety hyperproperties such as global robustness and fairness in LGNs, and we hope that it will serve as an inspiration for further research in this direction.

\medskip
\noindent\textbf{Related work.} 
We now briefly review related approaches to DNN verification for robustness. For a comprehensive overview, we refer to the survey paper of~\citet{DBLP:journals/ftopt/LiuALSBK21}. %
Most closely related to our approach are two recent confidence-based verifiers for 2-safety properties.
\citet{DBLP:conf/cav/AthavaleBCMNW24} encode global robustness and fairness for DNNs in SMT, whereas the concurrent work of Kabaha and Drachsler-Cohen~\citep{DBLP:journals/pacmpl/KabahaD24} targets global robustness of small image-classification DNNs using an MILP encoding.~\citet{DBLP:conf/date/WangHZ22} similarly encode the verification task as an MILP problem.  Further SMT-based approaches for fairness have also been explored by \citet{DBLP:conf/icse/BiswasR23} and \citet{DBLP:conf/aaai/KhedrS23}, but without incorporating a confidence-based verification framework. %
\citet{DBLP:conf/ijcai/RuanWSHKK19} propose a method for verifying global robustness by defining it as an
expectation of the maximal safe radius over a test
dataset. Different from them, in this paper, we focus on verifying logic gate neural networks. The verification of LGN-related networks has been addressed by~\citet{DBLP:conf/ijcai/BenamiraPYGH24}, but only for local robustness.

\section{Verifying Logic Gate Networks}\label{sec:verification}

In this section, we study the verification problem of logic gate networks. We begin by introducing the notations for LGNs, and then present our SAT-based encoding for global robustness and fairness.  

\subsection{Logic Gate Networks for Multi-Class Classification}

LGNs offer an alternative to conventional neural networks by replacing neurons with discrete logic gates. Formally, an LGN can be defined as a directed acyclic graph $\mathcal{G} = (\mathcal{V}, \mathcal{E})$ whose vertices \(\mathcal{V}\) are grouped into layers. Each vertex \(v \in \mathcal{V}\) in a layer corresponds to a two-input Boolean gate selected from a fixed set of 16 possible Boolean operations (e.g., NAND, XOR, etc.); the inputs are ordered since their positions affect non-commutative logic operations. The edges \(\mathcal{E}\) define connections from gates in one layer to gates in the next layer, resulting in a sparse, netlist-like architecture rather than dense matrix multiplications. Note that each gate can have multiple outgoing edges, allowing its output to be fed into several subsequent gates. The connectivity is randomly initialized and remains fixed throughout training. During training, each gate is modeled as a continuous relaxation over the possible Boolean operators, mixed via a softmax. LGNs are inherently non-linear and thus do not rely on non-linear activation functions. Prior work has demonstrated the ability of LGNs to learn tasks such as CIFAR-10 and MNIST~\citep{petersen2022deep,petersen2024convolutional}.

As in previous work, we employ LGNs for multi-class classification. Let \(\mathcal{X} \subseteq \mathbb{B}^d\) denote the input space (with inputs binarized using thermometer encoding for numerical features and one-hot encoding for categorical features), and let \(\mathcal{Y} = \{1,2,\dots,C\}\) be the set of class labels. We consider a classification model \( f: \mathcal{X} \to \mathcal{Y} \), where each input \( x \in \mathcal{X} \) is mapped to a predicted label \( f(x) \in \mathcal{Y} \). For $C$-class classification, $f(x)$ is obtained by an LGN's final layer producing $O$ Boolean outputs, partitioned into $C$ blocks of size $L$, where $O=C\times L$. Let $o_{j,k} \in \{0,1\}$ denote the $k$-th output bit for class $j$. As in \citet{petersen2022deep}:
$\text{score}(j) \;=\; \sum_{k=1}^{L} o_{j,k}$, for $j = 1,\dots,C$. The predicted label is the class with the highest score:
$f(x) \;=\; \arg\max_{\,1\leq j\leq C} \text{score}(j).$

We define the confidence for \( f(x) \):
\begin{equation}
\text{conf}(f(x)) = \frac{\text{score}(f(x))}{\sum_{j=1}^C \text{score}(j)}.
\label{eq:confidence}
\end{equation}

\subsection{Global Fairness and Robustness}
We define our notions of global fairness and robustness using a confidence-based approach inspired by \citet{DBLP:conf/cav/AthavaleBCMNW24}. %
Our setting assumes that inputs are discretized and encoded as binary vectors. This discretization leads to slight modifications in our definitions. To capture the aforementioned properties, we consider pairs of inputs, denoted \(x\) and \(x'\). Comparing two inputs allows us to formalize the idea that if the network predicts with high confidence on one input, then any sufficiently similar input should receive the same label. This is also referred to as self-composition \citep{barthe2011secure}. For a given confidence threshold \(\kappa > 0\) and a numerical tolerance \(\epsilon\), given as an integer, we require that
\begin{equation}
\forall x,\, x'\;:\quad \Phi(x,x',\epsilon) \wedge \text{conf}(f(x)) > \kappa \quad \Longrightarrow \quad f(x) = f(x'),
\label{eq:robustness}
\end{equation}
where the definition of the similarity condition \(\Phi(x,x',\epsilon)\) differs for global fairness and robustness.

\paragraph{Global Fairness.}  
For fairness, we partition the input \(x\) into sensitive attributes 
and non-sensitive attributes. Subscripts are used to denote the $i$th bit of the input $x$.
In this case, we require that the non-sensitive part satisfies the similarity condition, while the sensitive attributes differ. Formally, we define
\begin{equation}
\Phi(x,x',\epsilon) \;:=\; \Biggl( \bigwedge_{i \in \mathcal{N}_n} \, d(x_i,x'_i) \le \epsilon \Biggr) \wedge \Biggl( \bigwedge_{j \in \mathcal{C}_n} \, (x_j = x'_j) \Biggr) \wedge \Biggl( \bigwedge_{k \in \mathcal{C}_s} \, (x_k \neq x'_k) \Biggr),
\label{eq:fairness}
\end{equation}
where \(\mathcal{N}_n\) and \(\mathcal{C}_n\) denote non-sensitive numerical and categorical features, respectively, and \(\mathcal{C}_s\) indexes sensitive attributes. For numerical features encoded using thermometer encoding, the distance function \(d(\cdot,\cdot)\) counts the number of differing bits.

\paragraph{Global Robustness.}  
For global robustness, we require that each numerical feature differs by at most \(\epsilon\) and that the categorical features remain identical. Therefore, we can encode it like global fairness, with the number of sensitive attributes $\mathcal{C}_s$ set to the empty set.

These formulations, which capture conditions over pairs of inputs, are examples of {hyperproperties} \citep{clarkson2010hyperproperties}. More specifically, they are 2-safety properties. In the next section, we detail a SAT encoding to verify these properties for LGNs.
\subsection{SAT Encoding}
In this section, we present a SAT-based encoding for verifying global robustness and fairness defined above. As LGNs are fully discretized and operate on Boolean inputs and outputs, we can encode both the network behavior and the property constraints as Boolean formulas. In the following, we assume standard notations from classical Boolean logic. A literal is either a Boolean variable $v$ or its negation $\neg v$. A formula in conjunctive normal form (CNF) is a conjunction of clauses, with each clause being a disjunction of literals. We write $\texttt{f}(V)$ to denote a formula over sets of variables $V$ and $U,V$ denotes $U\cup V$. An assignment is a function that maps each variable to a Boolean value in $\{\bot,\top\}$. The satisfiability problem (SAT) is to determine, for a given CNF formula, whether there exists an assignment under which the formula is true. For clarity, in the following we use ``$\simeq$'' to denote syntactic equivalence between sets of variables, and ``$\Rightarrow$'' for semantic implication. We write $V[k]$ to denote the $k$-th variable in $V$.

First of all, we need to encode the network. Let \(V_{in}\) be the set of Boolean variables encoding an input \(x\in\mathcal{X}\) and \(V_{out}\) be the output Boolean variables representing \(f(x)\). Each logic gate in the LGN is encoded by its corresponding Boolean clause, so that the network is represented as: 
$V_{out} \simeq \texttt{network}(V_{in}).$
For a second input \(x'\), we define \(V'_{out} \) and \(V'_{in}\) symmetrically.
\subsubsection{Constraints on Inputs}
\paragraph{Well-Formedness.}  
We require that the assignments to \(V_{in}\) and \(V'_{in}\) conform to our encoding: numerical features must be correctly encoded using thermometer encoding, and categorical features must be one-hot encoded. We refer to this condition as \(\texttt{well\_formed}(V_{in})\) (and similarly for \(V'_{in}\)).

\paragraph{Numerical Proximity.}  
For each numerical feature \(i\) (with \(i\in \mathcal{N}\)), let \(V_{in}^{(i)}\) and \(V_{in}^{\prime (i)}\) denote the thermometer encodings (each with \(B\) bits). We enforce that the two encodings differ by at most \(\epsilon\) bit-flips. 
Formally, we define the proximity constraint for feature \(i\) as
\begin{equation}
\texttt{prox}_\epsilon\bigl(V_{in}^{(i)}, V_{in}^{\prime (i)}\bigr) \;:=\; 
\bigwedge_{k=\epsilon}^{B-1} \Bigl[ \Bigl( V_{in}^{(i)}[k] \Rightarrow V_{in}^{\prime (i)}[k-\epsilon] \Bigr) 
\wedge \Bigl( V_{in}^{\prime (i)}[k] \Rightarrow V_{in}^{(i)}[k-\epsilon] \Bigr) \Bigr].
\label{eq:prox}
\end{equation}
This ensures that the thermometer encodings for feature \(i\) in both inputs are within \(\epsilon\) bit-flips.

\paragraph{Categorical.} We enforce that categorical features remain identical across both inputs. This means that for each categorical feature \(i\), every bit in \(V^{(i)}_{in}\) and \(V'^{(i)}_{in}\) must be the same. We refer to this condition as \(\texttt{same\_cat}(V^{(i)}_{in}, V'^{(i)}_{in})\).

\subsubsection{Constraints on Outputs}

The output variables \(V_{out}\) are partitioned into \(C\) blocks, one for each class. Each block consists of a total of \(L\) bits. For class \(c\) (with \(c\in\{1,\dots,C\}\)), let \(V_{out}^{(c)}\) denote its corresponding block. To facilitate comparisons, we sort each block using a sorting network \citep{DBLP:journals/jsat/EenS06}:
\begin{equation}
V_{out}^{(c),\text{sorted}} \simeq \texttt{sort\_net}\Bigl(V_{out}^{(c)}\Bigr).
\label{eq:sortnet}
\end{equation}
The same procedure is applied to the corresponding block in the second output \(V'_{out}\). When the superscript is omitted, we refer to sorting all output bits in order to encode the overall sum of ones.

\paragraph{Winning Condition.}  
For each class \(c\), we define a Boolean variable \(w_c\) that encodes the condition that class \(c\) is the winning (predicted) class. Specifically, we require that:
\begin{equation}
w_c \;\Rightarrow\; \Biggl(
\underbrace{\bigwedge_{d < c} \; \bigwedge_{k=0}^{L-1} \Bigl( V_{out}^{(d),\text{sorted}}[k] \Rightarrow V_{out}^{(c),\text{sorted}}[k] \Bigr)}_{\text{class } c \text{ has confidence at least as high as classes } d < c} \;\wedge\;
\underbrace{\bigwedge_{d > c} \; \bigvee_{k=0}^{L-1} \Bigl( V_{out}^{(c),\text{sorted}}[k] \wedge \neg V_{out}^{(d),\text{sorted}}[k] \Bigr)}_{\text{class } c \text{ has confidence higher than classes } d > c}
\Biggr)
\label{eq:winning}
\end{equation}
Similarly, we define \(w'_c\) for the primed output \(V'_{out}\). Then, to capture the constraint that two inputs must be assigned different classes, we encode:
\begin{equation}
\texttt{diff\_class}(V_{out},V'_{out}) \;\coloneqq\; \bigwedge_{c=1}^{C} \Bigl( w_c \;\Rightarrow\; \neg w'_c \Bigr).
\label{eq:diffclass}
\end{equation}

\paragraph{Confidence.}  Finally, we define \(\texttt{confidence}_{>\kappa}(V)\) to require that the network’s output for input \(V\) exceeds a specified confidence threshold \(\kappa\). We can write:

\begin{equation}
\texttt{confidence}_{>\kappa}(V_{out}) \;:=\; \bigwedge_{i=1}^{C \cdot L} \left( V^{\text{sorted}}_{out}[i-1] \Rightarrow \bigvee_{c=1}^{C} \; V_{out}^{(c),\text{sorted}}\!\left[\left\lfloor i \cdot \kappa \right\rfloor\right] \right).
\label{eq:confidence_cnf}
\end{equation}

\subsubsection{Overall Verification Condition} 
To prove the validity of a propositional formula, we check whether its negation is unsatisfiable.
We first define an overall formula:
\begin{equation}
\begin{aligned}
\Psi \;:=\; & \texttt{well\_formed}(V_{in}) \land \texttt{well\_formed}(V'_{in}) \land\; \texttt{network}(V_{in}) \land \texttt{network}(V'_{in}) \\
            & \land\; \texttt{confidence}_{>\kappa}(V_{out}) \land\; \,\texttt{diff\_class}(V_{out},V'_{out}).
\end{aligned}
\label{eq:psi}
\end{equation}

We then specialize \(\Psi\) for our two properties as follows. First, \(\Psi_{\text{fair}}\) augments \(\Psi\) with the requirement that non-sensitive features are similar (numerically within \(\epsilon\) and categorically identical), while the sensitive features differ. Formally,
\begin{equation}
\begin{aligned}
\Psi_{\text{fair}} \;:= \; \Psi \land \Biggl( \Bigl( \bigwedge_{i \in N_n} \texttt{prox}_\epsilon\bigl(V_{in}^{(i)},V_{in}^{\prime(i)}\bigr)
\wedge \bigwedge_{j \in C_n} \texttt{same\_cat}\bigl(V_{in}^{(j)},V_{in}^{\prime(j)}\bigr) \Bigr) \\
 \wedge \;\Bigl( \bigwedge_{k \in \mathcal{C}_s} \lnot\,\texttt{same\_cat}\bigl(V_{in}^{(k)},V_{in}^{\prime(k)}\bigr) \Bigr) \Biggr).
\end{aligned}
\label{eq:psi_fair}
\end{equation}

Similarly, for \(\Psi_{\text{robust}}\), we use the same encoding as for \(\Psi_{\text{fair}}\), with the set of sensitive attributes \(\mathcal{C}_s\) being empty. Verification then reduces to proving that \(\Psi_{\text{robust}}\) or \(\Psi_{\text{fair}}\) is unsatisfiable. %

\section{Experiments}
\label{sec:experiments}
In this section, we present experimental results demonstrating the effectiveness of our proposed method for fairness and robustness verification of LGNs. We evaluate our approach on a range of classification tasks derived from five different datasets. 

\subsection{Datasets and Model Training Setup}

\paragraph{Datasets.}
In Table~\ref{tab:data-stats}, we summarize the key attributes of each dataset (number of instances, features, and classes). Four of these datasets—German Credit, Adult, Law School, and COMPAS—were used by prior work on DNN verification for global fairness and robustness verification \citep{DBLP:conf/cav/AthavaleBCMNW24, DBLP:conf/icse/BiswasR23}. 
We also investigate a five-class variant of the Adult dataset by partitioning annual income into five brackets, utilizing the Folktables dataset with data from the 2018 California census \citep{flood2021integrated}. %

\begin{table}[t]
\centering
\caption{Dataset (after preprocessing) showing the dataset size, total number of features including the number of numeric features and the number of output classes.}

\label{tab:data-stats}
\begin{tabular}{lccc}
\toprule
\textbf{Dataset} & \textbf{\#Samples} & \textbf{\#Features} (Categorical/Numeric) & \textbf{\#Classes} \\
\midrule
German Credit  \citep{hofmann1994german}   & 1,000   & 16 (12/4) & 2\\
Adult \citep{adult_2}            & 46,033  & 7 (4/3) & 2\\
Law School   \citep{wightman1998lsac}     & 21,982 & 9 (5/4) & 2\\
COMPAS \citep{Larson2017compas}           & 60,798  & 8 (7/1)  & 3\\
Adult-5 Class \citep{flood2021integrated}     & 195,665 & 7 (4/3)& 5 \\
\bottomrule
\end{tabular}
\end{table}

\paragraph{Data Splits and Encodings.}
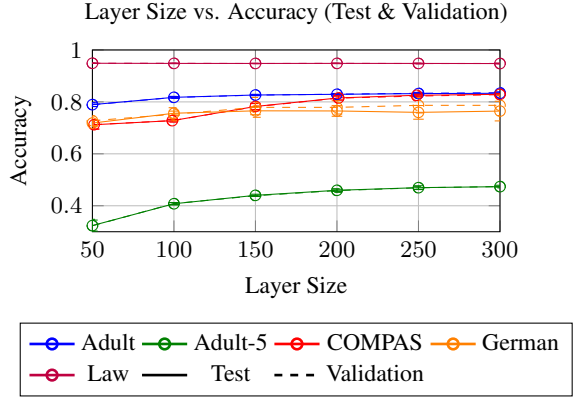
\begin{wrapfigure}[18]{r}{0.5\linewidth} %
    \centering\vspace{-1.5em}
    \pgfplotsset{
    discard if not/.style 2 args={
        x filter/.code={
            \edef\tempa{\thisrow{#1}}
            \edef\tempb{#2}
            \ifx\tempa\tempb
            \else
                \def\pgfmathresult{inf}
            \fi
        }
    }
}

\makeatletter
\pgfplotstableset{
    discard if not/.style 2 args={
        row predicate/.code={
            \def\pgfplotstable@loc@TMPd{\pgfplotstablegetelem{##1}{#1}\of}
            \expandafter\pgfplotstable@loc@TMPd\pgfplotstablename
            \edef\tempa{\pgfplotsretval}
            \edef\tempb{#2}
            \ifx\tempa\tempb
            \else
                \pgfplotstableuserowfalse
            \fi
        }
    }
}
\makeatother

\begin{tikzpicture}
\begin{axis}[
    font=\footnotesize,
    width=7.0cm,
    height=4cm,
    xlabel={Layer Size},
    ylabel={Accuracy},
    ymax=1, ymin=0.3,
    xmax=300, xmin=50,
    title={Layer Size vs. Accuracy (Test \& Validation)},
    grid=both,
    legend style={
        at={(0.5,-0.5)},
        anchor=north,
        font=\tiny,
        legend columns=4, %
        cells={align=center},
    },
    every axis legend/.append style={font=\footnotesize},
]

\addplot+[
    discard if not={model}{adult},
    forget plot,
    color=blue,
    mark=o,
    error bars/.cd,
        y dir=both,
        y explicit,
]
table[
    col sep=comma,
    trim cells=true,
    x=size,
    y=mean_test_acc,
    y error=std_test_acc,
]{data/accuracy_data.csv};
\addplot+[
    discard if not={model}{adult},
    color=blue,
    forget plot,
    mark=o,
    dashed,
    error bars/.cd,
        y dir=both,
        y explicit,
]
table[
    col sep=comma,
    trim cells=true,
    x=size,
    y=mean_val_acc,
    y error=std_val_acc,
]{data/accuracy_data.csv};

\addplot+[
    discard if not={model}{folktable},
    color=green!50!black,
    forget plot,
    mark=o,
    error bars/.cd,
        y dir=both,
        y explicit,
]
table[
    col sep=comma,
    trim cells=true,
    x=size,
    y=mean_test_acc,
    y error=std_test_acc,
]{data/accuracy_data.csv};

\addplot+[
    discard if not={model}{folktable},
    color=green!50!black,
    mark=o,
    forget plot,
    dashed,
    error bars/.cd,
        y dir=both,
        y explicit,
]
table[
    col sep=comma,
    trim cells=true,
    x=size,
    y=mean_val_acc,
    y error=std_val_acc,
]{data/accuracy_data.csv};

\addplot+[
    discard if not={model}{compas},
    color=red,
    forget plot,
    mark=o,
    error bars/.cd,
        y dir=both,
        y explicit,
]
table[
    col sep=comma,
    trim cells=true,
    x=size,
    y=mean_test_acc,
    y error=std_test_acc,
]{data/accuracy_data.csv};

\addplot+[
    discard if not={model}{compas},
    color=red,
    forget plot,
    mark=o,
    dashed,
    error bars/.cd,
        y dir=both,
        y explicit,
]
table[
    col sep=comma,
    trim cells=true,
    x=size,
    y=mean_val_acc,
    y error=std_val_acc,
]{data/accuracy_data.csv};

\addplot+[
    discard if not={model}{german},
    color=orange,
    mark=o,
    forget plot,
    solid,
    error bars/.cd,
        y dir=both,
        y explicit,
]
table[
    col sep=comma,
    trim cells=true,
    x=size,
    y=mean_test_acc,
    y error=std_test_acc,
]{data/accuracy_data.csv};

\addplot+[
    discard if not={model}{german},
    color=orange,
    mark=o,
    forget plot,
    dashed,
    error bars/.cd,
        y dir=both,
        y explicit,
]
table[
    col sep=comma,
    trim cells=true,
    x=size,
    y=mean_val_acc,
    y error=std_val_acc,
]{data/accuracy_data.csv};

\addplot+[
    discard if not={model}{lawNoWeights},
    color=purple,
    mark=o,
    forget plot,
    solid,
    error bars/.cd,
        y dir=both,
        y explicit,
]
table[
    col sep=comma,
    trim cells=true,
    x=size,
    y=mean_test_acc,
    y error=std_test_acc,
]{data/accuracy_data.csv};

\addplot+[
    discard if not={model}{lawNoWeights},
    color=purple,
    mark=o,
    forget plot,
    dashed,
    error bars/.cd,
        y dir=both,
        y explicit,
]
table[
    col sep=comma,
    trim cells=true,
    x=size,
    y=mean_val_acc,
    y error=std_val_acc,
]{data/accuracy_data.csv};
\addlegendimage{mark=o, color=blue, solid, line width=1pt}
\addlegendentry{Adult}
\addlegendimage{mark=o, color=green!50!black, solid, line width=1pt}
\addlegendentry{Adult-5}
\addlegendimage{mark=o, color=red, solid, line width=1pt}
\addlegendentry{COMPAS}
\addlegendimage{mark=o, color=orange, solid, line width=1pt}
\addlegendentry{German}
\addlegendimage{mark=o, color=purple, solid, line width=1pt}
\addlegendentry{Law}

\addlegendimage{mark=none, color=black, solid, line width=1pt}
\addlegendentry{Test}
\addlegendimage{mark=none, color=black, dashed, line width=1pt}
\addlegendentry{Validation}

\end{axis}

\end{tikzpicture}
    \caption{Validation and test accuracy achieved for five different classification tasks as a function of model layer size. Each point represents the mean accuracy over five runs with different random seeds. Error bars indicate standard deviation. }
    \label{fig:accuracy models}
\end{wrapfigure}

All datasets are split into 64\% training, 16\% validation, and 20\% test sets. Numeric attributes are discretized via thermometer encoding into a maximum of 20 buckets (except for German Credit, for which we use only 5 buckets due to a larger number of features). If the range of valid values is less than the maximum, fewer buckets are employed accordingly. Categorical features are one-hot encoded.
\paragraph{LGN Model Training.}

We train LGNs with three layers, each containing one of $\{50,100,150,200,250,300\}$ gates to investigate different model sizes and the impact on verification runtime. Figure \ref{fig:accuracy models} illustrates the validation and test accuracy of our models, when hardened, across different layer sizes. For COMPAS, we adjust the layer sizes to the closest number divisible by 3 to accommodate ternary output blocks. For each layer size and dataset, we run five different random seeds (shuffling both data splits and initial gate parameters). 
We use the Adam optimizer with a constant learning rate of 0.01 for 200 epochs. %

\subsection{Verification Framework}
In the experimental evaluation, we verify global fairness (GF) and global robustness (GR)—of LGN networks, as described in Section~\ref{sec:verification}. %
In all experiments, we use the award-winning SAT solver Kissat~\citep{BiereFallerFazekasFleuryFroleyksPollitt-SAT-Competition-2024-solvers} as the backend solver. All experiments are run on an AMD EPYC 9654 96-Core Processor.
We conduct two types of experiments:
\begin{enumerate}
    \item \textbf{Fixed-threshold experiments:} %
    We perform queries %
    at specific confidence thresholds (e.g., \(\kappa=0.5\), \(\kappa=0.99\)) for GF. In these experiments, we measure the solver’s runtime and record its outcome (UNSAT/SAT), corresponding to being globally fair or not, respectively.
    \item \textbf{Binary search over \(\kappa\):} For both GF and GR, we perform a binary search over \(\kappa\) (with a convergence threshold of 0.05) to identify the smallest safe confidence threshold at which no counterexample pair \((x,x')\) exists. For each trial \(\kappa\), we query the SAT solver. We then report the cumulative runtime across all queries as the total solving time.%
\end{enumerate}

\subsection{Global Fairness Verification}
\label{subsec:comparison_prior}
For global fairness, we fix the allowed perturbation \(\epsilon=0\) on the non-sensitive attributes, meaning that only the sensitive attributes (gender, ethnicity, or age – for German Credit) may vary.

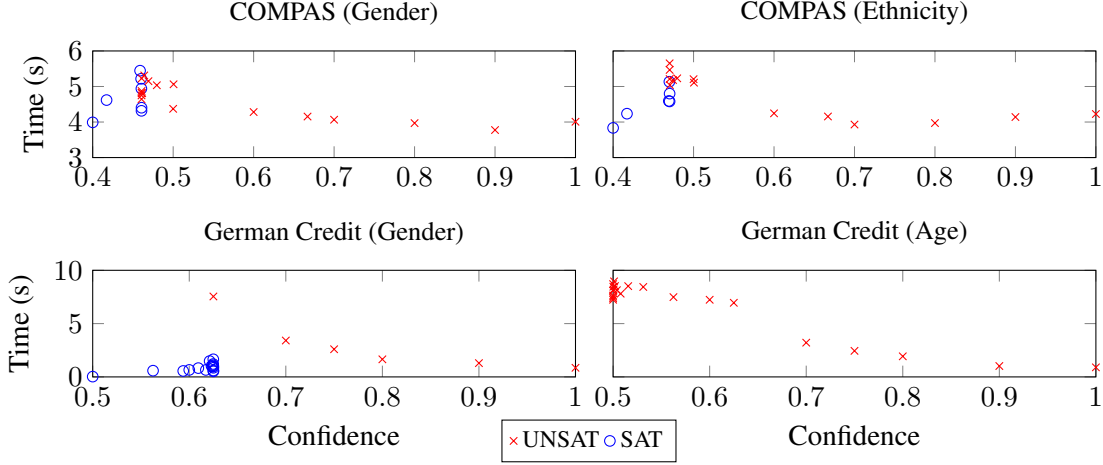
\begin{figure}[t]
    \centering
    
    \pgfplotsset{
    discard if not/.style 2 args={
        x filter/.code={
            \edef\tempa{\thisrow{#1}}
            \edef\tempb{#2}
            \ifx\tempa\tempb
            \else
                \def\pgfmathresult{inf}
            \fi
        }
    }
}

\makeatletter
\pgfplotstableset{
    discard if not/.style 2 args={
        row predicate/.code={
            \def\pgfplotstable@loc@TMPd{\pgfplotstablegetelem{##1}{#1}\of}
            \expandafter\pgfplotstable@loc@TMPd\pgfplotstablename
            \edef\tempa{\pgfplotsretval}
            \edef\tempb{#2}
            \ifx\tempa\tempb
            \else
                \pgfplotstableuserowfalse
            \fi
        }
    }
}
\makeatother

\begin{tikzpicture}
\begin{groupplot}[
  group style={group size=2 by 2, horizontal sep=0.5cm, vertical sep=1.5cm},
  xlabel={Confidence},
  ylabel={Time (s)},
  legend style={font=\footnotesize},
  title style={font=\small},
  ymin=3, ymax=6,
  xmin=0.4, xmax=1.0,
  width=8cm,         %
  height=3cm,        %
  legend style={
    at={(-0.05,-0.4)},
    anchor=north,
    legend columns=2,
    font=\footnotesize,
    },
  every axis legend/.append style={font=\footnotesize},
  ytick={3,4,5,6},
]

\nextgroupplot[title={COMPAS (Gender)}, xlabel={}]
\addplot+[only marks, mark=o, blue, discard if not={robustness}{False}] 
  table[col sep=comma]{300_compas_gender.csv};
\addplot+[only marks, mark=x, red, discard if not={robustness}{True}] 
  table[col sep=comma]{300_compas_gender.csv};

\nextgroupplot[title={COMPAS (Ethnicity)}, xlabel={}, ylabel={},yticklabels={}]
\addplot+[only marks, mark=o, blue, discard if not={robustness}{False}] 
  table[col sep=comma]{300_compas_race.csv};
\addplot+[only marks, mark=x, red, discard if not={robustness}{True}] 
  table[col sep=comma]{300_compas_race.csv};

\nextgroupplot[title={German Credit (Gender)},
            xmin=0.5, xmax=1.0,
            ymin=0, ymax=10,
            ytick={0,5,10},]
\addplot+[only marks, mark=o, blue, discard if not={robustness}{False}] 
  table[col sep=comma]{50_german_gender.csv};
\addplot+[only marks, mark=x, red, discard if not={robustness}{True}] 
  table[col sep=comma]{50_german_gender.csv};

\nextgroupplot[title={German Credit (Age)}, xlabel={Confidence},
            xmin=0.5, xmax=1.0,
            ymin=0, ymax=10,ylabel={},yticklabels={},
            ytick={0,5,10},]
\addplot+[only marks, mark=o, blue, discard if not={robustness}{False}] 
  table[col sep=comma]{50_german_age.csv};
\addplot+[only marks, mark=x, red, discard if not={robustness}{True}] 
  table[col sep=comma]{50_german_age.csv};
\addlegendimage{only marks, mark=o, blue}
\addlegendentry{UNSAT}
\addlegendimage{only marks, mark=x, red}
\addlegendentry{SAT}
\end{groupplot}
\end{tikzpicture}
    \caption{Solving time and result vs. confidence value for two example models (COMPAS with 300 and German Credit with 50 gates per layer). } %
    \label{fig:sat-solving-time-thresholds}
\end{figure}

\paragraph{Solving time at fixed \(\kappa\).} Figure \ref{fig:sat-solving-time-thresholds} shows solving times for one of our largest COMPAS models with 300 gates per layer and one of our smallest German Credit models with 50 gates per layer. We observe that solving time increase as we approach the globally robust boundary. %

All verification queries for these two models are performed in under 10 seconds and, except for German Credit with the sensitive attribute age, which is globally robust for any confidence threshold, we find non-trivial confidence thresholds. We additionally verified that at least one input example \(x\) exists that results in a higher confidence output than the found boundary.

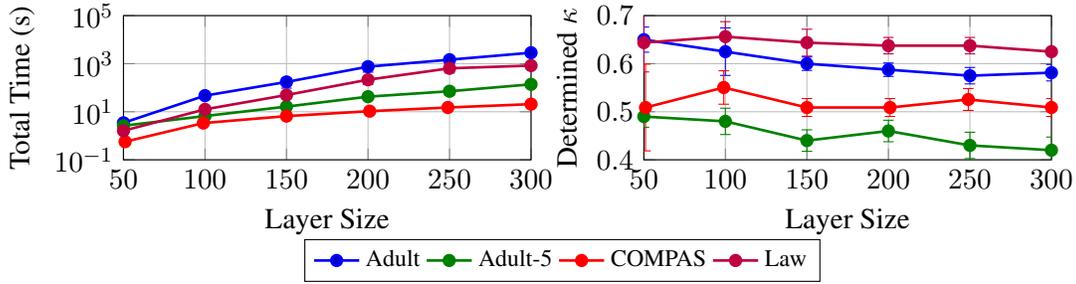
\begin{figure}[t]
    \centering
    \newcommand{\myAxisScaling}{%
    xmin=50, xmax=300,%
    ymin=0.1, ymax=100000,%
    ytick={0.1,10,1000,100000},
    }
    \def\myCSV{data/data_race_latex.csv}
    \pgfplotsset{
    discard if not/.style 2 args={
        x filter/.code={
            \edef\tempa{\thisrow{#1}}
            \edef\tempb{#2}
            \ifx\tempa\tempb
            \else
                \def\pgfmathresult{inf}
            \fi
        }
    }
}

\makeatletter
\pgfplotstableset{
    discard if not/.style 2 args={
        row predicate/.code={
            \def\pgfplotstable@loc@TMPd{\pgfplotstablegetelem{##1}{#1}\of}
            \expandafter\pgfplotstable@loc@TMPd\pgfplotstablename
            \edef\tempa{\pgfplotsretval}
            \edef\tempb{#2}
            \ifx\tempa\tempb
            \else
                \pgfplotstableuserowfalse
            \fi
        }
    }
}
\makeatother

\begin{tikzpicture}
\begin{groupplot}[
    group style={
        group size=2 by 1,
        horizontal sep=1.5cm,
    },
    width=7cm,
    height=3.5cm,
    grid=both,
    xlabel={Layer Size},
    legend style={
        at={(-0.2,-0.55)},
        anchor=north,
        legend columns=7,
        font=\footnotesize,
    },
    every axis legend/.append style={font=\footnotesize},
    every axis plot/.append style={line width=1pt},
]

\nextgroupplot[
    title={},
    ylabel={Total Time (s)},
    ymode=log,
    \myAxisScaling
]

\addplot+[
    discard if not={model}{adult},
    mark=*,
    color=blue,
    error bars/.cd,
        y dir=both,
        y explicit,
]
table[
    col sep=comma,
    trim cells=true,
    x=size,
    y=mean_total_time,
    y error=std_total_time,
]{\myCSV};

\addplot+[
    discard if not={model}{folktable},
    mark=*,
    color=green!50!black,
    mark options={fill=green!50!black},
    error bars/.cd,
        y dir=both,
        y explicit,
]
table[
    col sep=comma,
    trim cells=true,
    x=size,
    y=mean_total_time,
    y error=std_total_time,
]{\myCSV};

\addplot+[
    discard if not={model}{compas},
    color=red,
    mark=*,
        mark options={fill=red},
    error bars/.cd,
        y dir=both,
        y explicit,
]
table[
    col sep=comma,
    trim cells=true,
    x=size,
    y=mean_total_time,
    y error=std_total_time,
]{\myCSV};

\addplot+[
    discard if not={model}{lawNoWeights},
    mark=*,
    color=purple,
    error bars/.cd,
        y dir=both,
        y explicit,
]
table[
    col sep=comma,
    trim cells=true,
    x=size,
    y=mean_total_time,
    y error=std_total_time,
]{\myCSV};

\IfStrEq{\myCSV}{data/data_gender_latex.csv}{%
    \addplot+[
        discard if not={model}{german}, %
        color=orange,
        mark=*,
        mark options={color=orange},
        error bars/.cd,
            y dir=both,
            y explicit,
    ]
    table[
        col sep=comma,
        trim cells=true,
        x=size,
        y=mean_total_time,
        y error=std_total_time,
    ]{\myCSV};
}{}

\nextgroupplot[
    title={},
    ylabel={Determined \(\kappa\)},
    xmin=50,
    xmax=300,
    ymin=0.4, ymax=0.7,
    minor y tick num=0,
    ytick={0.4,0.5,0.6,0.7},
]
\addplot+[
    discard if not={model}{adult},
    color=blue,
    mark=*,
    error bars/.cd,
        y dir=both,
        y explicit,
]
table[
    col sep=comma,
    trim cells=true,
    x=size,
    y=mean_confidence,
    y error=std_confidence,
]{\myCSV};

\addplot+[
    discard if not={model}{folktable},
    color=green!50!black,
    mark=*,
    mark options={fill=green!50!black},
    error bars/.cd,
        y dir=both,
        y explicit,
]
table[
    col sep=comma,
    trim cells=true,
    x=size,
    y=mean_confidence,
    y error=std_confidence,
]{\myCSV};

\addplot+[
    discard if not={model}{compas},
    color=red,
    mark=*,
    mark options={fill=red},
    error bars/.cd,
        y dir=both,
        y explicit,
]
table[
    col sep=comma,
    trim cells=true,
    x=size,
    y=mean_confidence,
    y error=std_confidence,
]{\myCSV};

\addplot+[
    discard if not={model}{lawNoWeights},
    color=purple,
    mark=*,
    error bars/.cd,
        y dir=both,
        y explicit,
]
table[
    col sep=comma,
    trim cells=true,
    x=size,
    y=mean_confidence,
    y error=std_confidence,
]{\myCSV};

\IfStrEq{\myCSV}{latex-template/data/data_gender_latex.csv}{%
    \addplot+[
        discard if not={model}{german}, %
        color=orange,
        mark options={color=orange},
        mark=*,
        error bars/.cd,
            y dir=both,
            y explicit,
    ]
    table[
        col sep=comma,
        trim cells=true,
        x=size,
        y=mean_confidence,
        y error=std_confidence,
    ]{\myCSV};
}{}

\addlegendimage{mark=*, color=blue, solid, line width=1pt}
\addlegendentry{Adult}
\addlegendimage{mark=*, color=green!50!black, solid, line width=1pt}
\addlegendentry{Adult-5}
\addlegendimage{mark=*, color=red, solid, line width=1pt}
\addlegendentry{COMPAS}
\addlegendimage{mark=*, color=orange, solid, line width=1pt}
\addlegendentry{Law}
\IfStrEq{\myCSV}{latex-template/data/data_gender_latex.csv}{
    \addlegendimage{mark=*, color=orange, solid, line width=1pt}
    \addlegendentry{German}
}{}

\end{groupplot}
\end{tikzpicture}
    \caption{Comparison of runtime and determined confidence thresholds for the sensitive attribute ethnicity, as the number of gates per layer increases (standard deviation too small to see).}
    \label{fig:sat_solving_ethnicity}
\end{figure}

\paragraph{Minimum Confidence for Global Fairness.}
For the sensitive attributes gender and ethnicity, we perform a binary search over the confidence threshold \(\kappa\) to determine the smallest value at which the model is globally fair. We set the timeout to 8 hours, and all our models, except German Credit for layer sizes above \(100\), were successfully verified. In addition, for each found confidence threshold, we confirm that at least one input \(x\) that attains the corresponding confidence value exits, ensuring the threshold is not trivially satisfied.
Figure~\ref{fig:sat_solving_ethnicity} presents the cumulative runtime of the binary search (on a logarithmic scale) and the corresponding minimal confidence thresholds for the sensitive attribute ethnicity; results for gender are provided in the Appendix.

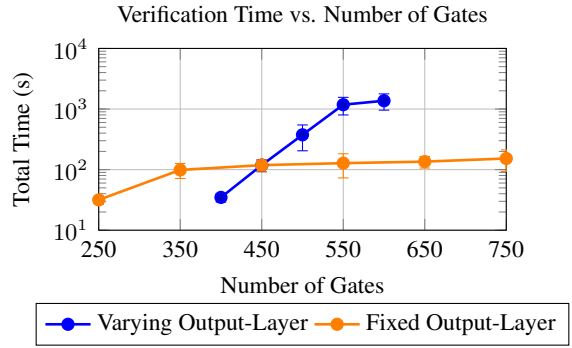
\begin{wrapfigure}[16]{r}{0.51\linewidth}
    \centering
    \pgfplotsset{
    discard if not/.style 2 args={
        x filter/.code={
            \edef\tempa{\thisrow{#1}}
            \edef\tempb{#2}
            \ifx\tempa\tempb
            \else
                \def\pgfmathresult{inf}
            \fi
        }
    }
}

\makeatletter
\pgfplotstableset{
    discard if not/.style 2 args={
        row predicate/.code={
            \def\pgfplotstable@loc@TMPd{\pgfplotstablegetelem{##1}{#1}\of}
            \expandafter\pgfplotstable@loc@TMPd\pgfplotstablename
            \edef\tempa{\pgfplotsretval}
            \edef\tempb{#2}
            \ifx\tempa\tempb
            \else
                \pgfplotstableuserowfalse
            \fi
        }
    }
}
\makeatother

\begin{tikzpicture}
\begin{axis}[
    font=\footnotesize,
    width=7.0cm,
    height=4cm,
    xlabel={Number of Gates},
    ylabel={Total Time (s)},
    xmin=250,xmax=750,
    title={Verification Time vs. Number of Gates},
    grid=both,
    legend style={
        at={(0.5,-0.4)},
        anchor=north,
        font=\footnotesize,
        legend columns=4, %
        cells={align=center},
    },
    every axis legend/.append style={font=\footnotesize},
    ymode=log,
    every axis plot/.append style={line width=1pt},
    minor x tick num=0,
    ytick={10,100,1000,10000},
    ymin=10,ymax=10000,
    minor grid style={draw=none},
    xtick={250,350,450,550,650,750},
    xticklabels={250,350,450,550,650,750},
]

\addplot+[
    forget plot,
    mark=*,
    color=blue,
    error bars/.cd,
        y dir=both,
        y explicit,
]
table[
    col sep=comma,
    trim cells=true,
    x=total_gates,
    y=mean_total_time,
    y error=std_total_time,
]{data_noskip_latex_gender_fixed_size.csv};

\addplot+[
    forget plot,
    color=orange,
    mark=*,
    mark options={fill=orange},
    error bars/.cd,
        y dir=both,
        y explicit,
]
table[
    col sep=comma,
    trim cells=true,
    x=total_gates,
    y=mean_total_time,
    y error=std_total_time,
]{data_noskip_latex_gender_fixed_output.csv};

\addlegendimage{ color=blue,mark=*, solid, line width=1pt}
\addlegendentry{Varying Output-Layer}
\addlegendimage{color=orange,mark=*, solid, line width=1pt}
\addlegendentry{Fixed Output-Layer}

\end{axis}

\end{tikzpicture}
    \caption{Adult dataset, varying vs. constant output layer size for 5 seeds. Either the output layer size or the size of the other layers is kept constant at $150$ gates.}
    \label{fig:diff_confidence_vs_fixed}
\end{wrapfigure}
Our results suggest that two main factors influence the solver's performance. %
First, the number of output classes appears to affect runtime, with more classes generally leading to faster verification, likely due to a coarser resolution in the confidence computation.  Second, the size of the input encoding—particularly the number of numerical features— impacts solving time. Although the Law School dataset has four numerical features, its values are represented using fewer buckets because the features have a limited range of integer values, which helps mitigate the expected slowdown. Further investigations on the effects of runtime are necessary in future work. 

For all datasets the found confidence values tend to stay constant or decrease with increasing model capacity. However, this is not the case for the sensitive attribute gender, see Appendix.
 \paragraph{Solving Time vs. Confidence Granularity.} 
As the confidence score is computed as the ratio of activated gates in the winning class to the total number of activated gates, its granularity is limited by the number of output gates. For example, in a Adult configuration with 300 gates (i.e., 150 gates per output class), the highest attainable non-1.0 confidence is \(\frac{150}{151} \approx 0.99\). Figure~\ref{fig:diff_confidence_vs_fixed} compares two experimental settings: one in which the output layer size is varied while all preceding layers are fixed at 150 gates, and another in which the output layer size is held constant at 150 gates while the sizes of the preceding layers are varied. It can be observed that increasing the output layer size leads to a substantial increase in verification time, whereas maintaining a constant output layer size results in only a small increase. 
With respect to the models used in Figure \ref{fig:diff_confidence_vs_fixed}, scaling the output layer yields a greater accuracy boost than adding gates earlier: the largest fixed output layer achieves  \(0.826 \pm 0.001\), versus \(0.834 \pm 0.003\) when the final layer’s gates are expanded to 300.

\subsection{Global Robustness Verification}

\begin{figure}[t]
    \centering
    \pgfplotsset{
    discard if not/.style 2 args={
        x filter/.code={
            \edef\tempa{\thisrow{#1}}
            \edef\tempb{#2}
            \ifx\tempa\tempb
            \else
                \def\pgfmathresult{inf}
            \fi
        }
    }
}

\makeatletter
\pgfplotstableset{
    discard if not/.style 2 args={
        row predicate/.code={
            \def\pgfplotstable@loc@TMPd{\pgfplotstablegetelem{##1}{#1}\of}
            \expandafter\pgfplotstable@loc@TMPd\pgfplotstablename
            \edef\tempa{\pgfplotsretval}
            \edef\tempb{#2}
            \ifx\tempa\tempb
            \else
                \pgfplotstableuserowfalse
            \fi
        }
    }
}
\newcommand{\epsonemarkershape}[1]{%
  \ifnum#1=1
    triangle*%
  \else
    \ifnum#1=4
      triangle*%
    \else
      \ifnum#1=5
        triangle*%
       \else
        *%
    \fi
    \fi
  \fi
}

\newcommand{\epsonemarkeroptions}[1]{%
  \ifnum#1=1
    0%
  \else
    180%
  \fi
}

\newcommand{\epscolor}[1]{%
  \ifnum#1=1
    red%
  \else
    \ifnum#1=2
      blue%
    \else
      \ifnum#1=3
        green!50!black%
      \else
        \ifnum#1=4
          purple%
        \else
          orange%
        \fi
      \fi
    \fi
  \fi
}

\newcommand{\epsonemarkershapetwo}[1]{%
  \ifnum#1=1
    *%
  \else
    *%
  \fi
}

\newcommand{\rotateiffourorfive}[1]{%
  \ifnum#1=4 180%
  \else\ifnum#1=5 180%
  \else 0%
  \fi\fi%
}

\makeatother
\def\myCSV{data/data_eps_adult.csv}
\def\myCSVfolklore{data/data_eps_folklore.csv}

\begin{tikzpicture}
\begin{groupplot}[
    group style={
        group size=2 by 2,
        horizontal sep=1.5cm,
        vertical sep=0.5cm,
    },
    width=7cm,
    height=3.5cm,
    grid=both,
    xlabel={Layer Size},
    legend to name=commonlegend,
    legend style={
        legend columns=5,
        font=\footnotesize,
    },
    every axis legend/.append style={font=\footnotesize},
    every axis plot/.append style={line width=1pt},
]

\nextgroupplot[
    title={},
    ylabel={\shortstack{\textbf{Adult}\\ \\ Total Time (s)}},
    ymode=log,
    ymin=0.1, ymax=100000,%
    ytick={0.1,10,1000,100000},
    xlabel={},
    xticklabels={},
    grid=both,
    minor x tick num=0,
    xmin=50, xmax=300,
]

\pgfplotsinvokeforeach{1,2,3,4,5} {
    \addplot+[
        discard if not={epsilon}{#1},
        mark=\epsonemarkershape{#1},
        mark options={rotate=\rotateiffourorfive{#1}, color=\epscolor{#1}},
        color=\epscolor{#1},
        solid,
        error bars/.cd,
            y dir=both,
            y explicit,
    ]
    table[
        col sep=comma,
        trim cells=true,
        x=size,
        y=mean_total_time,
        y error=std_total_time,
    ]{\myCSV};
}

\nextgroupplot[
    title={},
    ylabel={Determined $\kappa$},
    xlabel={},
    xticklabels={},
    ymin=0.5, ymax=1.0,
    grid=both,
    minor x tick num=0,
    minor y tick num=0,
    xmin=50, xmax=300,
    ytick={0.5,0.6,0.7,0.8,0.9,1.0},
]
\pgfplotsinvokeforeach{1,2,3,4,5} {
    \addplot+[
        discard if not={epsilon}{#1},
        mark=\epsonemarkershape{#1},
        mark options={rotate=\rotateiffourorfive{#1}, color=\epscolor{#1}},
        color=\epscolor{#1},
        solid,
        error bars/.cd,
            y dir=both,
            y explicit,
    ]
    table[
        col sep=comma,
        trim cells=true,
        x=size,
        y=mean_confidence,
        y error=std_confidence,
    ]{\myCSV};
}

\nextgroupplot[
    title={},
    ylabel={\shortstack{\textbf{Adult-5}\\ \\ Total Time (s)}},
    ymode=log,
    xmin=50, xmax=300,%
    ymin=0.1, ymax=100000,%
    ytick={0.1,10,1000,100000},
    grid=both,
    minor x tick num=0,
    minor y tick num=0,
    xmin=50, xmax=300,
]

\pgfplotsinvokeforeach{1,2,3,4,5} {
    \addplot+[
        discard if not={epsilon}{#1},
        mark=\epsonemarkershapetwo{#1},
       mark options={rotate=\rotateiffourorfive{#1}, color=\epscolor{#1}},
        color=\epscolor{#1},
        solid,
        error bars/.cd,
            y dir=both,
            y explicit,
    ]
    table[
        col sep=comma,
        trim cells=true,
        x=size,
        y=mean_total_time,
        y error=std_total_time,
    ]{\myCSVfolklore};
}

\nextgroupplot[
    title={},
    ylabel={Determined $\kappa$},
    ymin=0.5, ymax=1.0,
    grid=both,
    minor x tick num=0,
    minor y tick num=0,
    xmin=50, xmax=300,
    ytick={0.5,0.6,0.7,0.8,0.9,1.0},
]
\pgfplotsinvokeforeach{1,2,3,4,5} {
    \addplot+[
        discard if not={epsilon}{#1},
        mark=\epsonemarkershapetwo{#1},
     mark options={rotate=\rotateiffourorfive{#1}, color=\epscolor{#1}},
        color=\epscolor{#1},
        solid,
        error bars/.cd,
            y dir=both,
            y explicit,
    ]
    table[
        col sep=comma,
        trim cells=true,
        x=size,
        y=mean_confidence,
        y error=std_confidence,
    ]{\myCSVfolklore};
}

\addlegendimage{blue, solid, mark=\epsonemarkershape{1}, mark options={\epsonemarkeroptions{1}}, line width=1pt}
\addlegendentry{$\epsilon=1$}
\addlegendimage{red, solid, mark=\epsonemarkershape{2}, mark options={\epsonemarkeroptions{2}}, line width=1pt}
\addlegendentry{$\epsilon=2$}
\addlegendimage{green!50!black, solid, mark=\epsonemarkershape{3}, mark options={\epsonemarkeroptions{3}}, line width=1pt}
\addlegendentry{$\epsilon=3$}
\addlegendimage{purple, solid, mark=\epsonemarkershape{4}, mark options={\epsonemarkeroptions{4}}, line width=1pt}
\addlegendentry{$\epsilon=4$}
\addlegendimage{orange, solid, mark=\epsonemarkershape{5}, mark options={\epsonemarkeroptions{5}}, line width=1pt}
\addlegendentry{$\epsilon=5$}

\end{groupplot}

\node at (current bounding box.south) [anchor=north] {\pgfplotslegendfromname{commonlegend}};

\end{tikzpicture}
    \caption{Comparison of runtime and determined confidence values, as the number of gates per layer increases (standard deviation too small to see).}
    \label{fig:sat_solving_epsilon}
\end{figure}
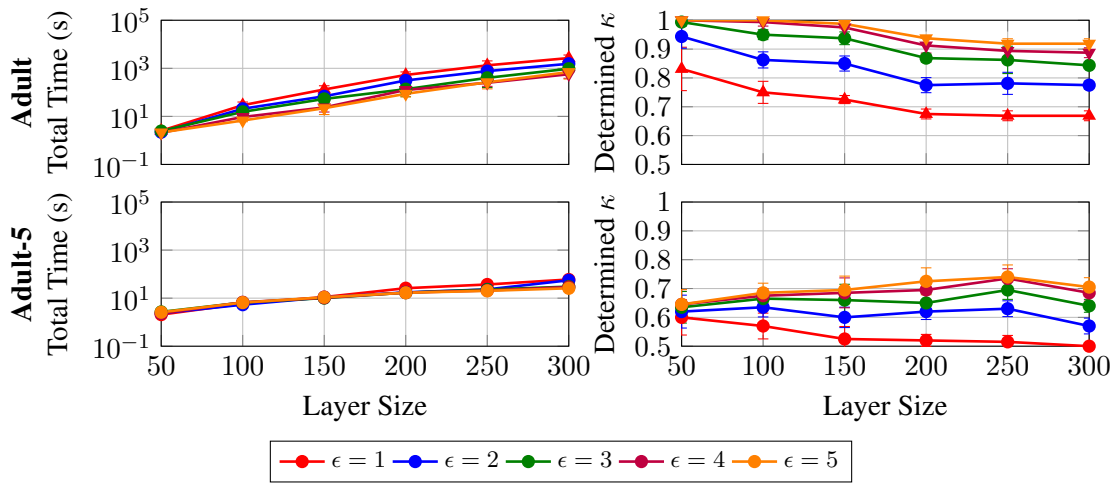

For global robustness, we allow up to \(\epsilon\) flips in each binarized numerical feature while keeping categorical features fixed. As with fairness, we perform a binary search over the confidence threshold \(\kappa\) and record the cumulative solver runtime for all queries. Figure~\ref{fig:sat_solving_epsilon} shows the returned minimal confidence thresholds and corresponding runtimes for various \(\epsilon\) values on the Adult and Adult-5 tasks. Results for other datasets are provided in the Appendix.
In Figure \ref{fig:sat_solving_epsilon}, we annotate each confidence threshold with markers indicating whether valid inputs \(x\) exist that yield the reported confidence. A filled upward triangle (\(\blacktriangle\)) denotes that all five model runs have at least one valid input resulting in the confidence value returned, a downward triangle (\(\blacktriangledown \)) indicates that no input \(x\) can produce the confidence value for any of the models, and a filled circle (\(\bullet\)) signifies that valid inputs exist for some, but not all, models. %
As expected, the runtime for \(\epsilon=1\) is generally the highest, which is a tighter constraint on the inputs compared with higher values of \(\epsilon\). For higher \(\epsilon\) values (e.g., \(\epsilon=4\) and \(\epsilon=5\)), no valid inputs exist for the Adult dataset; in contrast, for the Adult-5 class task, such inputs exist, at least for some of our models. %

\subsection{Comparison with previous work}
In general, direct comparisons are challenging as our approach employs LGNs and a SAT-based verification encoding, whereas the most closely related works \citet{DBLP:conf/cav/AthavaleBCMNW24} and \citet{DBLP:conf/icse/BiswasR23} verify conventional floating-point neural networks using SMT-based methods. 
\paragraph{Comparison with \citet{DBLP:conf/cav/AthavaleBCMNW24}.}
Even our smallest LGN models (with the exception of COMPAS, which requires models with more than 100 gates) achieve classification accuracies comparable to or \emph{higher than} those reported by \citet{DBLP:conf/cav/AthavaleBCMNW24}. Although they reported verification times for most instances under 60 seconds, our reproduction attempts sometimes yield runtimes on the \emph{order of minutes} (despite substantial communication with the authors, we could not reproduce their runtime numbers; possibly due to undocumented changes). Given these inconsistencies, we omit an experimental comparison.
In addition, their SMT-based verification relies on a softmax approximation that results in potentially spurious counterexamples. This limits the number of output classes due to increasing errors, whereas our method scales to a multi-class setting, as demonstrated by our 5-class variant of the Adult dataset. Lastly, \citet{DBLP:conf/cav/AthavaleBCMNW24} did not verify the existence of an input \(x\) that satisfies the minimum confidence threshold.

\paragraph{Comparison with \citet{DBLP:conf/icse/BiswasR23}.}
\citet{DBLP:conf/icse/BiswasR23} consider only binary classification and do not incorporate explicit confidence scores, hence operating at a confidence threshold of 0.5. Their investigated networks for the Adult and German Credit tasks yield similar accuracies to our LGNs. In contrast to our work, their experiments do not report any UNSAT instances, only observing time-out or SAT. Furthermore, they report that only one network can be verified with a straightforward SMT encoding (with a timeout of three days), necessitating the use of both sound and heuristic pruning schemes; the latter however may compromise accuracy. %

\section{Conclusion}

In this paper, we have taken a first step toward SAT-based verification of learned Logic Gate Networks (LGNs), focusing on the ease of verifying global fairness and robustness while retaining favorable predictive performance. Our initial experiments, including a 5-class dataset, show that LGNs can be verified efficiently using purely Boolean encodings, suggesting that such networks are well-suited for applications requiring strong correctness guarantees. 
Looking ahead, we plan to investigate more challenging datasets (e.g., complex image classification tasks) and tighten the integration between SAT solving and the verification tasks, for example, exploring the use of incremental SAT solving. In addition, we aim to explore real-world domains, such as neural network controllers for cyber-physical systems, where both efficient inference and reliable verification are critical. We hope this work will inspire further research into leveraging the inherently symbolic structure of LGNs for broader verification scenarios.

\acks{This work is supported in part by the ERC grant under Grant No. ERC-2020-AdG 101020093 and the Austrian Science Fund (FWF) [10.55776/COE12]. This research was supported by the Scientific Service Units (SSU) of ISTA through resources provided by Scientific Computing (SciComp).}

\bibliography{main}

 \newpage
 \begin{center}
 	\Large\textbf{Appendix}
 \end{center}

\begin{figure}[h!]
    \centering
    \newcommand{\myAxisScaling}{%
    xmin=50, xmax=300,%
    ymin=0.1, ymax=100000%
    }
    \def\myCSV{data/data_gender_latex.csv}
    \pgfplotsset{
    discard if not/.style 2 args={
        x filter/.code={
            \edef\tempa{\thisrow{#1}}
            \edef\tempb{#2}
            \ifx\tempa\tempb
            \else
                \def\pgfmathresult{inf}
            \fi
        }
    }
}

\makeatletter
\pgfplotstableset{
    discard if not/.style 2 args={
        row predicate/.code={
            \def\pgfplotstable@loc@TMPd{\pgfplotstablegetelem{##1}{#1}\of}
            \expandafter\pgfplotstable@loc@TMPd\pgfplotstablename
            \edef\tempa{\pgfplotsretval}
            \edef\tempb{#2}
            \ifx\tempa\tempb
            \else
                \pgfplotstableuserowfalse
            \fi
        }
    }
}
\makeatother

\begin{tikzpicture}
\begin{groupplot}[
    group style={
        group size=2 by 1,
        horizontal sep=1.5cm,
    },
    width=7cm,
    height=3.5cm,
    grid=both,
    xlabel={Layer Size},
    legend style={
        at={(-0.2,-0.55)},
        anchor=north,
        legend columns=7,
        font=\footnotesize,
    },
    every axis legend/.append style={font=\footnotesize},
    every axis plot/.append style={line width=1pt},
]

\nextgroupplot[
    title={},
    ylabel={Total Time (s)},
    ymode=log,
    \myAxisScaling
]

\addplot+[
    discard if not={model}{adult},
    mark=*,
    color=blue,
    error bars/.cd,
        y dir=both,
        y explicit,
]
table[
    col sep=comma,
    trim cells=true,
    x=size,
    y=mean_total_time,
    y error=std_total_time,
]{\myCSV};

\addplot+[
    discard if not={model}{folktable},
    mark=*,
    color=green!50!black,
    mark options={fill=green!50!black},
    error bars/.cd,
        y dir=both,
        y explicit,
]
table[
    col sep=comma,
    trim cells=true,
    x=size,
    y=mean_total_time,
    y error=std_total_time,
]{\myCSV};

\addplot+[
    discard if not={model}{compas},
    color=red,
    mark=*,
        mark options={fill=red},
    error bars/.cd,
        y dir=both,
        y explicit,
]
table[
    col sep=comma,
    trim cells=true,
    x=size,
    y=mean_total_time,
    y error=std_total_time,
]{\myCSV};

\addplot+[
    discard if not={model}{lawNoWeights},
    mark=*,
    color=purple,
    error bars/.cd,
        y dir=both,
        y explicit,
]
table[
    col sep=comma,
    trim cells=true,
    x=size,
    y=mean_total_time,
    y error=std_total_time,
]{\myCSV};

\IfStrEq{\myCSV}{data/data_gender_latex.csv}{%
    \addplot+[
        discard if not={model}{german}, %
        color=orange,
        mark=*,
        mark options={color=orange},
        error bars/.cd,
            y dir=both,
            y explicit,
    ]
    table[
        col sep=comma,
        trim cells=true,
        x=size,
        y=mean_total_time,
        y error=std_total_time,
    ]{\myCSV};
}{}

\nextgroupplot[
    title={},
    ylabel={Determined \(\kappa\)},
    xmin=50,
    xmax=300,
    ymin=0.4, ymax=0.7,
    minor y tick num=0,
    ytick={0.4,0.5,0.6,0.7},
]
\addplot+[
    discard if not={model}{adult},
    color=blue,
    mark=*,
    error bars/.cd,
        y dir=both,
        y explicit,
]
table[
    col sep=comma,
    trim cells=true,
    x=size,
    y=mean_confidence,
    y error=std_confidence,
]{\myCSV};

\addplot+[
    discard if not={model}{folktable},
    color=green!50!black,
    mark=*,
    mark options={fill=green!50!black},
    error bars/.cd,
        y dir=both,
        y explicit,
]
table[
    col sep=comma,
    trim cells=true,
    x=size,
    y=mean_confidence,
    y error=std_confidence,
]{\myCSV};

\addplot+[
    discard if not={model}{compas},
    color=red,
    mark=*,
    mark options={fill=red},
    error bars/.cd,
        y dir=both,
        y explicit,
]
table[
    col sep=comma,
    trim cells=true,
    x=size,
    y=mean_confidence,
    y error=std_confidence,
]{\myCSV};

\addplot+[
    discard if not={model}{lawNoWeights},
    color=purple,
    mark=*,
    error bars/.cd,
        y dir=both,
        y explicit,
]
table[
    col sep=comma,
    trim cells=true,
    x=size,
    y=mean_confidence,
    y error=std_confidence,
]{\myCSV};

\IfStrEq{\myCSV}{latex-template/data/data_gender_latex.csv}{%
    \addplot+[
        discard if not={model}{german}, %
        color=orange,
        mark options={color=orange},
        mark=*,
        error bars/.cd,
            y dir=both,
            y explicit,
    ]
    table[
        col sep=comma,
        trim cells=true,
        x=size,
        y=mean_confidence,
        y error=std_confidence,
    ]{\myCSV};
}{}

\addlegendimage{mark=*, color=blue, solid, line width=1pt}
\addlegendentry{Adult}
\addlegendimage{mark=*, color=green!50!black, solid, line width=1pt}
\addlegendentry{Adult-5}
\addlegendimage{mark=*, color=red, solid, line width=1pt}
\addlegendentry{COMPAS}
\addlegendimage{mark=*, color=orange, solid, line width=1pt}
\addlegendentry{Law}
\IfStrEq{\myCSV}{latex-template/data/data_gender_latex.csv}{
    \addlegendimage{mark=*, color=orange, solid, line width=1pt}
    \addlegendentry{German}
}{}

\end{groupplot}
\end{tikzpicture}
    \caption{Comparison of runtime and determined confidence thresholds for the sensitive attribute gender, as the number of gates per layer increases. For the German Credit task, only one model with 150 gates layer size could be verified within the set timeout of 8 hours. Larger German Credit models could not be verified within the timeout.}
    \label{fig:sat_solving_gender}
\end{figure}
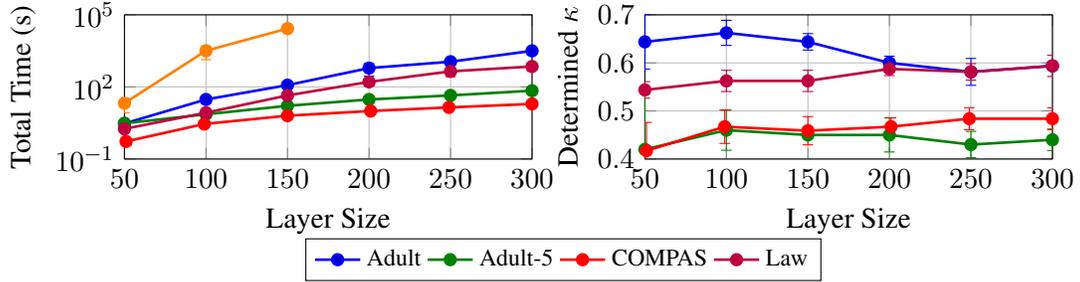

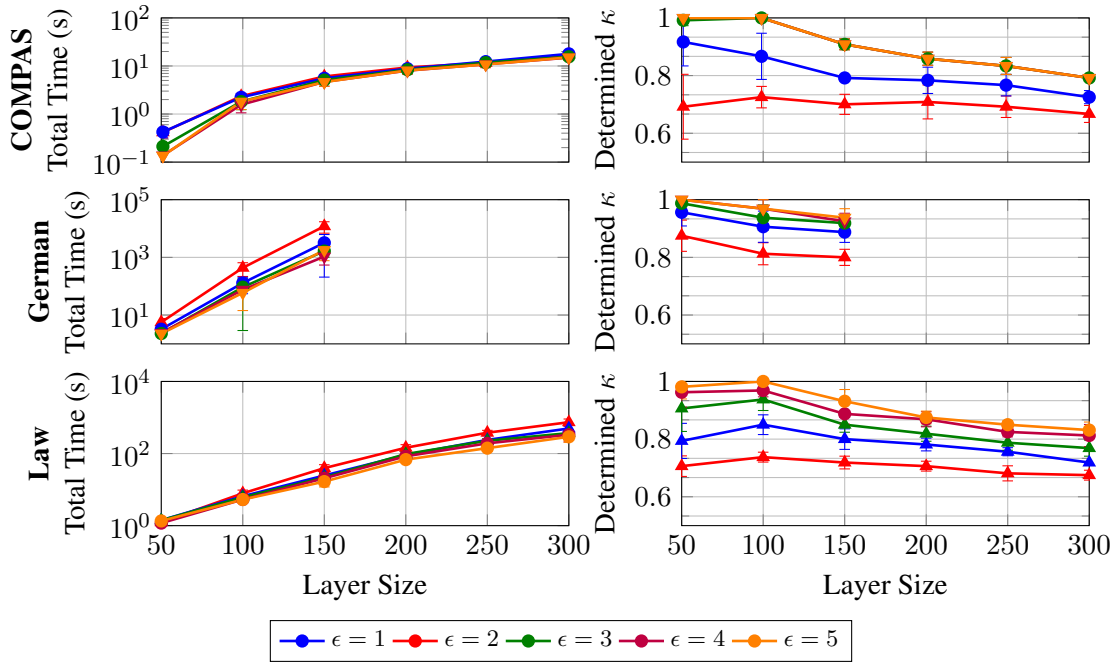
\begin{figure}[h!]
    \centering
    \pgfplotsset{
    discard if not/.style 2 args={
        x filter/.code={
            \edef\tempa{\thisrow{#1}}
            \edef\tempb{#2}
            \ifx\tempa\tempb
            \else
                \def\pgfmathresult{inf}
            \fi
        }
    }
}

\makeatletter
\pgfplotstableset{
    discard if not/.style 2 args={
        row predicate/.code={
            \def\pgfplotstable@loc@TMPd{\pgfplotstablegetelem{##1}{#1}\of}
            \expandafter\pgfplotstable@loc@TMPd\pgfplotstablename
            \edef\tempa{\pgfplotsretval}
            \edef\tempb{#2}
            \ifx\tempa\tempb
            \else
                \pgfplotstableuserowfalse
            \fi
        }
    }
}
\newcommand{\epsonemarkershape}[1]{%
  \ifnum#1=1
    triangle*%
  \else
    \ifnum#1=4
      triangle*%
    \else
      \ifnum#1=5
        triangle*%
       \else
        *%
    \fi
    \fi
  \fi
}

\newcommand{\epsonemarkershapelaw}[1]{%
  \ifnum#1=1
    triangle*%
  \else
    \ifnum#1=2
      triangle*%
    \else
      \ifnum#1=3
        triangle*%
       \else
        *%
    \fi
    \fi
  \fi
}

\newcommand{\epsonemarkeroptions}[1]{%
  \ifnum#1=1
    0%
  \else
    180%
  \fi
}

\newcommand{\epscolor}[1]{%
  \ifnum#1=1
    red%
  \else
    \ifnum#1=2
      blue%
    \else
      \ifnum#1=3
        green!50!black%
      \else
        \ifnum#1=4
          purple%
        \else
          orange%
        \fi
      \fi
    \fi
  \fi
}

\newcommand{\epsonemarkershapetwo}[1]{%
  \ifnum#1=1
    *%
  \else
    *%
  \fi
}

\newcommand{\rotateiffourorfive}[1]{%
  \ifnum#1=4 180%
  \else\ifnum#1=5 180%
  \else 0%
  \fi\fi%
}

\newcommand{\rotatecompas}[1]{%
  \ifnum#1=4 180%
  \else\ifnum#1=5 180%
  \else 0%
  \fi\fi%
}

\makeatother
\def\myCSVcompas{data/data_eps_compas.csv}
\def\myCSVgerman{data/data_eps_german_dropped_rows.csv}
\def\myCSVlaw{data/data_eps_law.csv}

\begin{tikzpicture}
\begin{groupplot}[
    group style={
        group size=2 by 3,
        horizontal sep=1.5cm,
        vertical sep=0.5cm,
    },
    width=7cm,
    height=3.5cm,
    grid=both,
    xlabel={Layer Size},
    legend to name=commonlegend,
    legend style={
        legend columns=5,
        font=\footnotesize,
    },
    every axis legend/.append style={font=\footnotesize},
    every axis plot/.append style={line width=1pt},
]

\nextgroupplot[
    title={},
    ylabel={\shortstack{\textbf{COMPAS}\\ \\ Total Time (s)}},
    ymode=log,
    ymin=0.1, ymax=100,
    xlabel={},
    xticklabels={},
    grid=both,
    minor x tick num=0,
    xmin=50, xmax=300,
    minor grid style={draw=none},
]

\pgfplotsinvokeforeach{1,2,3,4,5} {
    \addplot+[
        discard if not={epsilon}{#1},
        mark=\epsonemarkershape{#1},
        mark options={rotate=\rotateiffourorfive{#1}, color=\epscolor{#1}},
        color=\epscolor{#1},
        solid,
        error bars/.cd,
            y dir=both,
            y explicit,
    ]
    table[
        col sep=comma,
        trim cells=true,
        x=size,
        y=mean_total_time,
        y error=std_total_time,
    ]{\myCSVcompas};
}

\nextgroupplot[
    title={},
    ylabel={Determined $\kappa$},
    xlabel={},
    xticklabels={},
    ymin=0.5, ymax=1.0,
    grid=both,
    minor x tick num=0,
    minor y tick num=2,
    xmin=50, xmax=300,
]
\pgfplotsinvokeforeach{1,2,3,4,5} {
    \addplot+[
        discard if not={epsilon}{#1},
        mark=\epsonemarkershape{#1},
        mark options={rotate=\rotateiffourorfive{#1}, color=\epscolor{#1}},
        color=\epscolor{#1},
        solid,
        error bars/.cd,
            y dir=both,
            y explicit,
    ]
    table[
        col sep=comma,
        trim cells=true,
        x=size,
        y=mean_confidence,
        y error=std_confidence,
    ]{\myCSVcompas};
}

\nextgroupplot[
    title={},
    ylabel={\shortstack{\textbf{German}\\ \\ Total Time (s)}},
    xticklabels={},
    xlabel={},
    ymode=log,
    ymin=1, 
    ymax=100000,
    grid=both,
    minor x tick num=0,
    minor y tick num=0,
    xmin=50, xmax=300,
]

\pgfplotsinvokeforeach{1,2,3,4,5} {
    \addplot+[
        discard if not={epsilon}{#1},
        mark=\epsonemarkershape{#1},
       mark options={rotate=\rotateiffourorfive{#1}, color=\epscolor{#1}},
        color=\epscolor{#1},
        solid,
        error bars/.cd,
            y dir=both,
            y explicit,
    ]
    table[
        col sep=comma,
        trim cells=true,
        x=size,
        y=mean_total_time,
        y error=std_total_time,
    ]{\myCSVgerman};
}

\nextgroupplot[
    title={},
    xlabel={},
    xticklabels={},
    ylabel={Determined $\kappa$},
    ymin=0.5, ymax=1.0,
    grid=both,
    minor x tick num=0,
    minor y tick num=2,
    xmin=50, xmax=300,
]
\pgfplotsinvokeforeach{1,2,3,4,5} {
    \addplot+[
        discard if not={epsilon}{#1},
        mark=\epsonemarkershape{#1},
     mark options={rotate=\rotateiffourorfive{#1}, color=\epscolor{#1}},
        color=\epscolor{#1},
        solid,
        error bars/.cd,
            y dir=both,
            y explicit,
    ]
    table[
        col sep=comma,
        trim cells=true,
        x=size,
        y=mean_confidence,
        y error=std_confidence,
    ]{\myCSVgerman};
}

\nextgroupplot[
    title={},
    ylabel={\shortstack{\textbf{Law}\\ \\ Total Time (s)}},
    ymode=log,
    ymin=1, 
    ymax=10000,
    grid=both,
    minor x tick num=0,
    minor y tick num=0,
    xmin=50, xmax=300,
]

\pgfplotsinvokeforeach{1,2,3,4,5} {
    \addplot+[
        discard if not={epsilon}{#1},
        mark=\epsonemarkershapelaw{#1},
       mark options={rotate=\rotateiffourorfive{#1}, color=\epscolor{#1}},
        color=\epscolor{#1},
        solid,
        error bars/.cd,
            y dir=both,
            y explicit,
    ]
    table[
        col sep=comma,
        trim cells=true,
        x=size,
        y=mean_total_time,
        y error=std_total_time,
    ]{\myCSVlaw};
}

\nextgroupplot[
    title={},
    ylabel={Determined $\kappa$},
    ymin=0.5, ymax=1.0,
    grid=both,
    minor x tick num=0,
    minor y tick num=2,
    xmin=50, xmax=300,
]
\pgfplotsinvokeforeach{1,2,3,4,5} {
    \addplot+[
        forget plot,
        discard if not={epsilon}{#1},
        mark=\epsonemarkershapelaw{#1},
        mark options={rotate=\rotateiffourorfive{#1}, color=\epscolor{#1}},
        color=\epscolor{#1},
        solid,
        error bars/.cd,
            y dir=both,
            y explicit,
    ]
    table[
        col sep=comma,
        trim cells=true,
        x=size,
        y=mean_confidence,
        y error=std_confidence,
    ]{\myCSVlaw};
}

\addlegendimage{blue, solid, mark=*, line width=1pt}
\addlegendentry{$\epsilon=1$}
\addlegendimage{red, solid, mark=*, line width=1pt}
\addlegendentry{$\epsilon=2$}
\addlegendimage{green!50!black, solid, mark=*, mark options={}, line width=1pt}
\addlegendentry{$\epsilon=3$}
\addlegendimage{purple, solid, mark=*, mark options={}, line width=1pt}
\addlegendentry{$\epsilon=4$}
\addlegendimage{orange, solid, mark=*, mark options={}, line width=1pt}
\addlegendentry{$\epsilon=5$}

\end{groupplot}

\node at (current bounding box.south) [anchor=north] {\pgfplotslegendfromname{commonlegend}};

\end{tikzpicture}
    \caption{Comparison of runtime and determined confidence values for global robustness, as the number of gates per layer increases. Note the varying y-axis scaling on the logarithmic runtime plots. For German Credit, some higher epsilon values could be verified within the timeout time for some models with more than 150 gates. These are omitted from the Figure.}
    \label{fig:sat_solving_epsilon_appendix}
\end{figure}

\end{document}